\documentclass[11pt]{article}
\usepackage{url}
\usepackage[utf8]{inputenc}
\usepackage{amsmath,color,amssymb,amsthm,bbm}
\usepackage[letterpaper,margin=1.0in]{geometry}
\usepackage[ruled,linesnumbered]{algorithm2e}
\usepackage{array}
\usepackage{multirow}
\usepackage[hidelinks]{hyperref}
\usepackage{booktabs}
\usepackage{epsfig}
\usepackage{epstopdf}
\usepackage{algpseudocode}
\usepackage{graphicx}
\usepackage{color}
\usepackage{authblk}

\definecolor{officegreen}{rgb}{0.0, 0.5, 0.0}

\def\R{\mathbb{R}}

\def\W{\mathrm{W}}

\usepackage{bm}

\def\HK{\mathrm{HK}}
\def\KL{\textnormal{KL}}

\title{Neighbor Embeddings Using Unbalanced Optimal Transport Metrics}
\author[1]{Muhammad Rana}
\author[1,2]{Keaton Hamm}
\affil[1]{Department of Mathematics,  University of Texas at Arlington}
\affil[2]{Division of Data Science, University of Texas at Arlington}

\date{}

\begin{document}
%
\maketitle
\begin{abstract}
This paper proposes the use of the Hellinger--Kantorovich metric from unbalanced optimal transport (UOT) in a dimensionality reduction and learning (supervised and unsupervised) pipeline. The performance of UOT is compared to that of regular OT and Euclidean-based dimensionality reduction methods on several benchmark datasets including MedMNIST. The experimental results demonstrate that, on average, UOT shows improvement over both Euclidean and OT-based methods as verified by statistical hypothesis tests. In particular, on the MedMNIST datasets, UOT outperforms OT in classification 81\% of the time. For clustering MedMNIST, UOT outperforms OT 83\% of the time and outperforms both other metrics 58\% of the time.
\end{abstract}

\section{Introduction}

With advancements in sensing systems including those in medical imaging like MRI and CT scanners, the advent of the Internet of Things, and cheap abundant data storage, the volume of data collection has grown rapidly. Oftentimes, datasets in these fields contain spatial, temporal, and contextual features, and thus datasets have grown not only in volume, but also in dimensionality and complexity. High-dimensionality can lead to challenges for access, analysis, and interpretation. Without additional assumptions on structure within the data, there is typically some form of the curse of dimensionality that appears in a given task-oriented pipeline. For this reason, many works have explored when and how low-dimensional structures appear in high-dimensional data, and have considered how to detect and utilize such structure effectively for learning \cite{carlsson2009topology, bruna2013invariant}. 

One structure that is abundant in theory and practice is that of a low-dimensional embedded manifold \cite{meilua2024manifold}. Manifold learning  is an umbrella term to capture various algorithms that utilize the manifold structure to learn embeddings of the data into a lower-dimensional (typically Euclidean) space.   Examples from the zoo of manifold learning algorithms are Laplacian eigenmaps \cite{belkin2003laplacian}, diffusion maps \cite{coifman2006diffusion}, multidimensional scaling (MDS) \cite{ mardia1979multivariate}, t-stochastic neighbor embedding (t-SNE) \cite{maaten2008visualizing}, uniform manifold approximation and projection (UMAP) \cite{mcinnes2018umap}, and Isomap \cite{tenenbaum2000global}. All these but t-SNE are examples of global neighbor embeddings in the terminology of \cite{meilua2024manifold}, which work by forming a neighborhood ($\varepsilon$ or $k$-nearest neighbor) graph on the data, with edge weights determined by a metric on the ambient data space. Choosing $\varepsilon$ and $k$ must be balanced by the manifold geometry (e.g., curvature and reach), and sampling density of the data from the manifold must be higher for highly curved manifolds \cite{bernstein2000graph}.  

Vectorizing images and using Euclidean distance to describe the geometry lacks the ability to maintain local spatial information. To mitigate this issue, several works have proposed considering image data as probability measures on a grid in $\R^2$ \cite{hamm2023wassmap,kolouri2017optimal, wang2010optimal}, where the notion of distance between images becomes from the Wasserstein metric arising from optimal transport (OT) theory \cite{peyre2019computational}. Loosely speaking, the Wasserstein distance between two images corresponds to a minimal energy morphing of one image into another (see Section \ref{SEC:Background} for a formal definition). One reason for the benefit of using transport-based metrics is that they keep local spatial information. A few works have studied manifold learning when the Wasserstein distance is the ambient metric on the data space \cite{wang2010optimal,hamm2023wassmap,mathews2019molecular}. In \cite{wang2010optimal}, MDS with the Wasserstein distance is used for classification and compared to a feature-based approach for pathological image data. They found that OT performs similar or better compared to the feature-based approach. In \cite{mathews2019molecular}, the authors use the Wasserstein-1 distance (or Earth-mover's distance) with Diffusion maps to discover low-dimensional patterns in gene expression networks resulting in a new, more robust way of extracting features for genetic contributions to patients developing soft-tissue sarcoma.  

However, one of the limitations of using the Wasserstein metric is that it requires data to be probability measures, and so does not allow for variations in mass. But in many cases, mass naturally differs and carries meaning. The Hellinger--Kantorovich distances \cite{liero2016optimal} arise from considering unbalanced optimal transport (UOT), and are well-defined metrics on the space of nonnegative measures on a Euclidean domain. These distances allow for different total masses of the data measures by creating or annihilating mass. 

This work proposes the use of the UOT-based Hellinger--Kantorovich metrics when forming neighbor embeddings of image data to relax the rigidness of the Wasserstein metric and allow for more meaningful modeling of the data geometry to account for unequal mass and more complex image structures. We study this problem by studying classification and clustering task performance on embedded data with the use of three metrics: Euclidean, Wasserstein-2, and Hellinger--Kantorovich. We compute the accuracies over multiple trials, and compare results via statistical hypothesis test. We use recent medical imaging benchmark datasets from MedMNIST \cite{yang2023medmnist} to understand performance on diverse data as well as more classic datasets like MNIST and Coil-20. We find that UOT embeddings outperform OT embeddings a majority of the time for both clustering and classification across all datasets, and UOT embeddings yield better task performance even more often on the MedMNIST datasets.

\section{Background}\label{SEC:Background}



Let $\mathcal{P}_2(\R^n)$, be the set of all probability measures on $\R^n$ with finite $2$-nd moment: $\int_{\R^n}|x|^2d\mu(x)<\infty$. The Wasserstein-2 distance is defined by the Kantorovich problem
\begin{equation}\label{eq:W2-metric}
    \W_2(\mu,\nu) := \inf_{\pi\in\Gamma(\mu,\nu)} \left(\int_{\R^{n}\times\R^n}|x-y|^2d\pi(x,y)\right)^\frac12,
\end{equation}
    where $\Gamma(\mu,\nu):=\{\gamma\in\mathcal{P}(\R^{n}\times\R^n): \gamma(A\times \R^n) = \mu(A),\; \gamma(\R^n\times A)=\nu(A) \textnormal{ for all Borel measurable } A\subset\R^n\}$ are couplings, i.e., joint measures whose marginals are $\mu$ and $\nu$. For discrete measures, one can exactly compute an optimal coupling via a linear program \cite{peyre2019computational} or use entropic regularization to approximate the distance quickly \cite{cuturi2013sinkhorn}.
    
To allow for unequal mass, we will use the Hellinger--Kantorovich metric on $\mathcal{M}_+(\Omega)$, the space of nonnegative measures on a bounded convex domain $\Omega\subset\R^n$ \cite{liero2016optimal,bonafini2023hellinger}, which for a given length scale $\kappa>0$ is given by
\begin{equation}\label{EQN:HK}
\HK_\kappa^2(\mu,\nu):=\inf_{\gamma\in\mathcal{M}_+(\Omega\times\Omega)}\left\{\int_{\Omega\times \Omega}c_\kappa(x,y)d\gamma(x,y) +\KL(P_1\gamma\ |\ \mu)+\KL(P_2\gamma\ |\ \mu)\right\},
\end{equation}
where KL is the Kullback--Leibler divergence, $P_i \gamma$, $i=1,2$ denotes the $i$-th marginal of the coupling measure $\gamma$, and the cost $c_\kappa$ is given by
\begin{equation}\label{EQN:HKCost}
c_\kappa:= \begin{cases}
    -2\log\cos(|x-y|/\kappa), & |x-y|\leq\kappa\pi/2,\\
    +\infty, & \textnormal{otherwise.}
\end{cases}    
\end{equation}
The cost function is such that within a length of $\kappa\pi/2$, mass transportation is prioritized, whereas no mass is transported outside that scale, but is rather created and destroyed. 



Beyond accounting for mass difference, unbalanced OT can reveal surprisingly low–dimensional structures that are invisible in Euclidean space. For example, consider a dataset in $\mathcal{M}_+(\Omega)$, $\Omega\subset\R^2$, given by translated indicator functions of the unit disk $\{\mathbbm{1}_D(\cdot-t_i)\}_{i=1}^N$. As a subset of $(\mathcal{M}_+(\Omega),\HK_\kappa)$, this is a 2-dimensional manifold (when $\kappa$ is suitably large) that is nearly isometric to the translation set $\{t_i\}_{i=1}^N\subset\R^2$ via the MDS embedding (see \cite{hamm2023wassmap} for similar reasoning for $W_2$) due to the fact that $\HK_\kappa(\mathbbm{1}_D(\cdot-t_i),\mathbbm{1}_D(\cdot-t_j)) \approx c|t_i-t_j|$ for large $\kappa$. On the other hand, if these were objects imaged by a camera or similar imaging device by being mapped to an $n$-pixel grid, the collection of images $\{x_i\}_{i=1}^N\subset\R^n$ would be typically full rank (depending on the spacing of the translation set), and would not be concentrated near a 2-dimensional manifold. Examples like this suggest that modeling data using the HK geometry can yield a better low-dimensional model, and potentially lead to improved task performance, which is the main purpose of our study here.

\section{Datasets} We use 5 benchmark datasets for both classification and clustering experiments: MNIST, Coil-20 \cite{Nene1996}, and for more sophisticated image types, we use BloodMNIST, OrganCMNIST, and RetinaMNIST from the MedMNIST v2 benchmark \cite{yang2023medmnist}. These are more varied in terms of texture, morphology, and complexity compared to the basic benchmarks, but are nonetheless curated, so they provide a fruitful testing ground for the use of UOT metrics in neighbor embeddings.



Each of the datasets we use from MedMNIST are from the set of 2D images, all of which have resolution $28\times 28$ after preprocessing \cite{yang2023medmnist}. BloodMNIST is a collection of 17,092 preprocessed images of 8 types of blood cells (basophils,  eosinophils, erythroblasts, immature
granulocytes (promyelocytes, myelocytes, and metamyelocytes), lymphocytes, monocytes, neutrophils and platelets), where the classification task is to determine what type of cell the image contains. OrganCMNIST comprises 23,660 2D coronal plane images from 3D CT scans from 11 different organs, and the task is to determine which images belong to which organ. RetinaMNIST consists of $1,600$ images of human retinas with 4 different grades of retinopathy caused by diabetes, where the task is to classify the grade of retinopathy.

\section{Methodology} 
In all experiments, we utilize a manifold learning pipeline of the form data $\to$ neighbor embedding $\to$ task, where the tasks are classification and clustering. We form embeddings via MDS, Isomap, t-SNE, and Laplacian eigenmaps using Euclidean, $\W_2$, and $\HK_1$ metrics. We exactly compute $\W_2$ distances using the python optimal transport package \cite{flamary2021pot}, and use the code from \cite{cai2022linearized} to compute $\HK$ distances. 


For classification, the embedded data in the low-dimensional Euclidean space is split into $80\%$-$20\%$ training and testing data, and we use the following classification algorithms on the train-test split in the embedded space: Linear Discriminant Analysis (LDA), k-Nearest Neighbors (KNN) for $k = 1, 3, 5$, linear Support Vector Machine (SVM(L)), RBF kernel SVD (SVM(R)), Random Forest, and Multinomial Logistic Regression (MLR). To cluster embedded data, we use $k$-means and spectral clustering. To compute the clustering accuracy, linear sum assignment is used to match the ground truth labels with the cluster labels. 

  
To determine which embedding outperforms the others (if any), we use statistical hypothesis tests to compare the mean of 10 replicate experiments for each of the Euclidean, OT and UOT-based embeddings. This is done by carrying out 6 hypothesis tests formulated with \textbf{Null Hypothesis}: The task accuracy using the embedding metric $A$ is less than or equal to that using metric $B$, and  \textbf{Alternative Hypothesis:} The task accuracy using the embedding metric $A$ is better than that using metric $B$. We use a significance level $p = 0.05$ for all hypothesis tests, and run two one-sided t-tests for each pair $A,B \in \{$Euclidean, $\W_2$, $\HK_1\}$ (where the roles of $A$ and $B$ are reversed in the second test). 


The embedding dimension is a hyperparameter for neighbor embeddings, and for our experiments, we follow the heuristic of \cite{medina2019heuristic}, based on the fact that local Singular Value Decomposition (SVD) provides a good estimation for the dimension of a tangent plane to data in a small neighborhood, and use the SVD to compute the embedding dimension as the integer that accounts for a fixed proportion of the variance of the spectrum, i.e., choose the minimum integer $n$ so that
\begin{equation}\label{EQN:EmbDim}\frac{\sum_{i=1}^n \sigma_i^2}{\sum_{j=1}^N\sigma_j^2} \geq a,\end{equation}
for fixed $a\in(0,1)$. 

We used $1,000$ data points for each experiment for each dataset. We compute the embedding dimension via \eqref{EQN:EmbDim}, typically with the value of $a=.97$, however, for some datasets, this leads to a very small embedding dimension (e.g., 1), so for these we augment $a$ somewhat. For MNIST, Coil-20, and OrganCMNIST, we use $a=.97$ corresponding to embedding dimensions 129, 66, and 120, respectively. For BloodMNIST and RetinaMNIST, we use $a=.99$ and $a=.999$, respectively, corresponding to embedding dimensions 21 and 80. Note that in using the sklearn implementation of t-SNE, data is embedded into dimension 3 for all experiments.


\section{Clustering experiment results}

Table \ref{clustering results} reports the clustering accuracy with columns being datasets and rows being the clustering algorithms ($k$-means or spectral clustering) applied to data embedded by MDS, Isomap, t-SNE or Eigenmaps. Bold entries mean that the given metric outperformed all others at confidence level $p=.05$, and bars over Euclidean or UOT values mean they outperformed the other, but did not significantly outperform OT.

\begin{table*}[!ht]
\centering
\scriptsize
\begin{tabular}{|l|c|c|c|c|c|c|}
\hline 
Method & Algorithm & MNIST & Coil-20 & BloodMNIST & OrganCMNIST & RetinaMNIST \\
\hline 
 & & Euc\textbar OT\textbar UOT & Euc\textbar OT\textbar UOT & Euc\textbar OT\textbar UOT & Euc\textbar OT\textbar UOT & Euc\textbar OT\textbar UOT \\
\hline
\multirow{2}{*}{\textbf{MDS}} \rule{0pt}{1em}    & $k$-Means   & 0.51\textbar0.43\textbar0.49 & \textbf{0.64}\textbar0.50\textbar0.56 & 0.45\textbar0.33\textbar0.45 & \textbf{0.39}\textbar0.31\textbar0.36 & \textbf{0.34}\textbar0.28\textbar0.32   \\
                                     & Spectral & 0.60\textbar0.49\textbar\textbf{0.64} & 0.67\textbar0.68\textbar$\overline{0.69}$ & 0.48\textbar0.34\textbar\textbf{0.52} & 0.34\textbar0.31\textbar\textbf{0.38} & \textbf{0.36}\textbar0.29\textbar0.31 \\
\hline 
\multirow{2}{*}{\textbf{Isomap}} \rule{0pt}{1em}      & $k$-means   & 0.57\textbar0.59\textbar0.59 & \textbf{0.69}\textbar0.61\textbar0.65  & 0.45\textbar0.36\textbar0.43 & $\overline{0.34}$\textbar0.33\textbar0.29 & 0.30\textbar0.28\textbar\textbf{0.38}  \\
                                     & Spectral & \textbf{0.59}\textbar0.50\textbar0.58 & 0.52\textbar0.56\textbar\textbf{0.61}& \textbf{0.35}\textbar0.34\textbar0.30 & 0.38\textbar0.34\textbar\textbf{0.44} & 0.35\textbar0.29\textbar\textbf{0.40} \\
\hline 
\multirow{2}{*}{\textbf{t-SNE}} \rule{0pt}{1em}      & $k$-means   & 0.57\textbar0.62\textbar$\overline{0.62}$ & 0.72\textbar0.70\textbar\textbf{0.72} & 0.44\textbar0.37\textbar0.44 & 0.41\textbar0.38\textbar\textbf{0.45} & 0.30\textbar0.26\textbar\textbf{0.32}  \\
                                     & Spectral & 0.59\textbar\textbf{0.62}\textbar0.58 & \textbf{0.48}\textbar0.42\textbar0.41 & 0.42\textbar0.36\textbar\textbf{0.45}& 0.41\textbar0.36\textbar\textbf{0.45} & 0.42\textbar0.37\textbar0.42  \\
\hline  
\multirow{2}{*}{\textbf{Eigenmaps}} \rule{0pt}{1em}      & $k$-means   & 0.38\textbar0.49\textbar$\overline{0.49}$ & 0.44\textbar\textbf{0.65}\textbar$\overline{0.57}$ & 0.42\textbar0.32\textbar\textbf{0.46} & 0.36\textbar0.32\textbar\textbf{0.40} & 0.32\textbar0.32\textbar\textbf{0.37}  \\
                                     & Spectral & 0.38\textbar\textbf{0.49}\textbar$\overline{0.44}$ & $\overline{0.51}$\textbar0.52\textbar0.44 & 0.36\textbar0.31\textbar\textbf{0.41}& 0.37\textbar0.33\textbar\textbf{0.38} & 0.31\textbar\textbf{0.36}\textbar$\overline{0.33}$  \\
\hline 
\end{tabular}\caption{Clustering results for all datasets applying $k$-means and spectral clustering algorithms on data embedded via MDS, Isomap, t-SNE, or Laplacian Eigenmaps. In each cell, the metric is Euclidean (left), OT (middle) and UOT (right).}
\label{clustering results}
\end{table*}

Performance of most methods is relatively similar on MNIST and Coil-20, though UOT tends to perform better for spectral clustering on MDS embeddings. Additionally, UOT frequently outperforms OT in these datasets. Indeed, UOT outperforms OT 70\% (28/40) of the time. For the MedMNIST datasets, UOT outperforms the other metrics much of the time (58\% of the trials), and more interestingly, most of the time (83\% of trials) outperforms OT by a large magnitude. 

The MedMNIST clustering results seem promising in that UOT almost always outperforms OT. Given that these datasets involve various textures and morphologies, it seems that the ability of UOT to create and destroy mass allows for the manifold learning methods to find a more concrete structure, whereas for less texture-based data like MNIST and Coil-20, OT sometimes performs comparably to UOT.

\section{Classification experiment results}

Synthesizing the classification results in Tables \ref{classification_mnist_raw_wass_hk}--\ref{classification_medmnist_retinamnist_raw_hk_wass}, we see the following trends. For metric MDS embeddings, UOT typically outperforms both Euclidean and OT by a substantial margin, with an exception of RetinaMNIST. For Isomap embeddings, which are designed to capture global nonlinear manifolds, UOT typically outperforms both other metrics on OrganCMNIST and RetinaMNIST, while results are mixed for the rest of the datasets. That said, UOT typically performs better than OT on BloodMNIST. 

For t-SNE embeddings, the results vary quite a lot from dataset to dataset. In some instances Euclidean embeddings are better, including Coil-20, while the other datasets typically show no overall winner. However, oftentimes, the UOT results are still much better than the OT results. Eigenmap embeddings also yield varied outcomes, but notably the majority of the time, UOT performs as well or better than OT.

\begin{table}[!ht]
   \centering
  \scriptsize
  \begin{tabular}{|>{\hspace{-3pt}}l<{\hspace{-3pt}}|>{\hspace{-1pt}}c<{\hspace{-1pt}}|>{\hspace{-1pt}}c<{\hspace{-1pt}}|>{\hspace{-1pt}}c<{\hspace{-1pt}}|>{\hspace{-1pt}}c<{\hspace{-1pt}}|}
    \hline
      & MDS & Isomap & t-SNE & Eigenmaps \\ \hline
    Metric & Euc\textbar OT\textbar UOT & Euc\textbar OT\textbar UOT & Euc\textbar OT\textbar UOT & Euc\textbar OT\textbar UOT
    
    \\ \hline \rule{0pt}{1em}
    LDA & 0.81\textbar0.83\textbar0.79 & 
    0.84\textbar0.86\textbar0.86 & 
    0.70\textbar0.55\textbar0.68 & 
    0.87\textbar0.90\textbar\textbf{0.92} \\
    
    1NN & 0.84\textbar0.79\textbar\textbf{0.90} & 0.84\textbar0.86\textbar$\overline{0.87}$ & 0.87\textbar0.76\textbar0.88 & 
    0.69\textbar0.87\textbar$\overline{0.87}$ \\
    
    3NN & 0.83\textbar0.80\textbar\textbf{0.89} & 0.84\textbar0.87\textbar$\overline{0.86}$ & 0.87\textbar0.77\textbar0.88 & 
    0.56\textbar0.89\textbar$\overline{0.88}$ \\
    
    5NN & 0.84\textbar0.81\textbar\textbf{0.89} & 0.83\textbar0.86\textbar0.85 & 
    0.85\textbar0.76\textbar0.87 & 
    0.49\textbar0.89\textbar$\overline{0.88}$ \\
    
    SVM (L) & 0.86\textbar0.20\textbar\textbf{0.87} & 0.84\textbar0.83\textbar\textbf{0.88} &
    0.76\textbar0.63\textbar0.73 & 
    0.12\textbar0.12\textbar0.12 \\
    
    SVM (R) & 0.85\textbar0.90\textbar\textbf{0.90} & 0.87\textbar0.90\textbar$\overline{0.89}$ & \textbf{0.82}\textbar0.72\textbar0.80 & 0.85\textbar0.90\textbar$\overline{0.90}$ \\
    
    RF & 0.82\textbar0.82\textbar\textbf{0.88} & 0.83\textbar0.87\textbar$\overline{0.87}$ & 0.87\textbar0.75\textbar0.87 & 
    0.86\textbar0.88\textbar$\overline{0.89}$ \\
    
    MLR & 0.80\textbar0.41\textbar\textbf{0.82} & 0.83\textbar0.81\textbar0.85 & 
    \textbf{0.72}\textbar0.59\textbar0.68 & 
    0.12\textbar\textbf{0.24}\textbar$\overline{0.17}$ \\
    \hline
  \end{tabular} \caption{Classification results for MNIST on Euclidean (left), OT (middle) and UOT (right) neighbor embeddings.}
  \label{classification_mnist_raw_wass_hk}
\end{table}

\begin{table}[!ht]
\centering
\scriptsize
\begin{tabular}{|>{\hspace{-3pt}}l<{\hspace{-3pt}}|>{\hspace{-1pt}}c<{\hspace{-1pt}}|>{\hspace{-1pt}}c<{\hspace{-1pt}}|>{\hspace{-1pt}}c<{\hspace{-1pt}}|>{\hspace{-1pt}}c<{\hspace{-1pt}}|}
\hline
Classifier & MDS & Isomap & t-SNE & Eigenmaps \\
\hline
    Metric & Euc\textbar OT\textbar UOT & Euc\textbar OT\textbar UOT & Euc\textbar OT\textbar UOT & Euc\textbar OT\textbar UOT
    
    \\ \hline \rule{0pt}{1em}
LDA        & \textbf{0.93} \textbar0.77\textbar0.90

                   & 0.95\textbar0.95\textbar0.95
                   
                   & \textbf{0.79}\textbar0.74\textbar0.75
                   
                   & \textbf{0.98}\textbar0.10\textbar0.09 \\
                   
1NN        & 0.94\textbar0.90\textbar\textbf{0.98}

                   & 0.94\textbar\textbf{0.96}\textbar$\overline{0.96}$
                   
                   & \textbf{0.99}\textbar0.90\textbar0.90
                   
                   & \textbf{0.97}\textbar0.91\textbar0.89 \\
                   
3NN        & 0.92\textbar0.86\textbar\textbf{0.95}

                   & 0.89\textbar0.93\textbar$\overline{0.94}$
                   
                   & \textbf{0.97}\textbar0.88\textbar0.90
                   
                   & \textbf{0.92}\textbar0.89\textbar0.86\\
                   
5NN        & 0.89\textbar0.83\textbar\textbf{0.93}

                   & 0.86\textbar0.92\textbar$\overline{0.90}$
                   
                   & \textbf{0.95}\textbar0.86\textbar0.87
                   
                   & \textbf{0.89}\textbar0.86\textbar0.86 \\
                   
SVM(L)   & 0.97\textbar0.07\textbar\textbf{0.99}

                   & 0.99\textbar0.84\textbar0.99
                   
                   & 0.85\textbar0.82\textbar0.85
                   
                   & 0.04\textbar0.07\textbar0.05 \\
                   
SVM(R)   & \textbf{0.97}\textbar0.87\textbar0.95

                   & 0.91\textbar\textbf{0.94}\textbar0.91
                   
                   & \textbf{0.86}\textbar0.78\textbar0.80
                   
                   & \textbf{0.97}\textbar0.88\textbar0.87 \\
                   
RF        & 0.92\textbar0.87\textbar\textbf{0.98}

                   & 0.97\textbar0.97\textbar\textbf{0.98}
                   
                   & \textbf{0.98}\textbar0.90\textbar0.90
                   
                   & \textbf{0.97}\textbar0.92\textbar0.90 \\
                   
MLR        & 0.95\textbar0.24\textbar0.94

                   & 0.98\textbar0.84\textbar0.98
                   
                   & \textbf{0.80}\textbar0.74\textbar0.77
                   
                 & 0.04\textbar0.15\textbar$\overline{0.12}$\\
\hline
\end{tabular}\caption{Classification results for Coil-20 on Euclidean (left), OT (middle) and UOT (right) neighbor embeddings.}
\label{classification_coil20_raw_hk_wass}
\end{table}

\begin{table}[!ht]
\centering
\scriptsize
\begin{tabular}{|>{\hspace{-3pt}}l<{\hspace{-3pt}}|>{\hspace{-1pt}}c<{\hspace{-1pt}}|>{\hspace{-1pt}}c<{\hspace{-1pt}}|>{\hspace{-1pt}}c<{\hspace{-1pt}}|>{\hspace{-1pt}}c<{\hspace{-1pt}}|}
\hline
Classifier & MDS & Isomap & t-SNE & Eigenmaps \\
\hline 
    Metric & Euc\textbar OT\textbar UOT & Euc\textbar OT\textbar UOT & Euc\textbar OT\textbar UOT & Euc\textbar OT\textbar UOT
    
    \\ \hline \rule{0pt}{1em}
LDA & 0.62\textbar0.52\textbar0.62

                   & 0.59\textbar0.57\textbar\textbf{0.61}
                   
                   &0.54\textbar0.49\textbar0.56
                   
                   & 0.66\textbar0.57\textbar0.67 \\
                   
1NN& 0.54\textbar0.43\textbar\textbf{0.62}

                   & 0.54\textbar0.47\textbar\textbf{0.56}
                   
                   & 0.61\textbar0.46\textbar\textbf{0.65}
                   
                   & 0.55\textbar0.49\textbar\textbf{0.60} \\
                   
3NN& 0.53\textbar0.44\textbar\textbf{0.62}

                   & 0.56\textbar0.50\textbar0.57
                   
                   &0.63\textbar0.47\textbar0.64
                   
                   & 0.58\textbar0.50\textbar\textbf{0.62}\\
                   
5NN        & 0.55\textbar0.46\textbar\textbf{0.63}

                   & 0.58\textbar0.51\textbar0.58
                   
                   & 0.63\textbar0.48\textbar0.63
                   
                   & 0.60\textbar0.52\textbar\textbf{0.62} \\
                   
SVM(L)   & 0.63\textbar0.21\textbar0.63

                   & 0.58\textbar0.32\textbar0.58
                   
                   & 0.57\textbar0.51\textbar0.58
                   
                   & 0.21\textbar0.19\textbar0.20 \\
                   
SVM(R)   & 0.63\textbar0.56\textbar\textbf{0.72}

                   & 0.65\textbar0.58\textbar0.64
                   
                   & 0.59\textbar0.53\textbar0.61
                   
                   & 0.68\textbar0.56\textbar0.68 \\
                   
RF         & 0.62\textbar0.51\textbar\textbf{0.70}

                   & 0.63\textbar0.57\textbar0.63
                   
                   & 0.63\textbar0.50\textbar\textbf{0.65}
                   
                   & 0.66\textbar0.56\textbar0.67 \\
                   
MLR        & 0.64\textbar0.21\textbar0.65

                   & 0.58\textbar0.44\textbar\textbf{0.61}
                   
                   & 0.55\textbar0.50\textbar0.57
                   
                   & 0.21\textbar0.20\textbar0.22\\
\hline 
\end{tabular}\caption{Classification results for BloodMNIST on Euclidean (left), OT (middle) and UOT (right) neighbor embeddings.}
\label{classification_medmnist_bloodmnist_raw_hk_wass}
\end{table}

\begin{table}[!h]
\centering
\scriptsize
\begin{tabular}{|>{\hspace{-3pt}}l<{\hspace{-3pt}}|>{\hspace{-1pt}}c<{\hspace{-1pt}}|>{\hspace{-1pt}}c<{\hspace{-1pt}}|>{\hspace{-1pt}}c<{\hspace{-1pt}}|>{\hspace{-1pt}}c<{\hspace{-1pt}}|}
\hline
Classifier & MDS & Isomap & t-SNE & Eigenmaps \\
\hline 
    Metric & Euc\textbar OT\textbar UOT & Euc\textbar OT\textbar UOT & Euc\textbar OT\textbar UOT & Euc\textbar OT\textbar UOT
    
    \\ \hline \rule{0pt}{1em}
LDA & \textbf{0.67}\textbar0.55\textbar0.61

                   & 0.68\textbar0.61\textbar\textbf{0.71}
                   
                   & 0.48\textbar0.44\textbar\textbf{0.55}
                   
                   & 0.75\textbar0.66\textbar0.76 \\
                   
1NN        & 0.52\textbar0.53\textbar\textbf{0.75}

                   & 0.66\textbar0.65\textbar\textbf{0.71}
                   
                   & \textbf{0.75}\textbar0.53\textbar0.72
                   
                   & 0.65\textbar0.65\textbar\textbf{0.75} \\
                   
3NN        & 0.47\textbar0.52\textbar\textbf{0.70}

                   & 0.63\textbar0.62\textbar\textbf{0.70}
                   
                   & \textbf{0.70}\textbar0.52\textbar0.67
                   
                   & 0.53\textbar0.64\textbar\textbf{0.71}\\
                   
5NN        & 0.51\textbar0.53\textbar\textbf{0.70}

                   & 0.64\textbar0.62\textbar\textbf{0.71}
                   
                   & \textbf{0.70}\textbar0.53\textbar0.67
                   
                   & 0.47\textbar0.64\textbar\textbf{0.67} \\
                   
SVM(L)   & 0.63\textbar0.23\textbar\textbf{0.67}

                   & 0.68\textbar0.38\textbar\textbf{0.71}
                   
                   & 0.52\textbar0.44\textbar\textbf{0.59}
                   
                   & 0.25\textbar0.26\textbar0.25 \\
                   
SVM(R)   & 0.63\textbar 0.61\textbar\textbf{0.72}

                   & 0.71\textbar0.66\textbar0.71
                   
                   & 0.64\textbar0.50\textbar0.60
                   
                   & 0.72\textbar0.70\textbar\textbf{0.76} \\
                   
RF         & 0.66\textbar0.60\textbar\textbf{0.76}

                   & 0.69\textbar0.64\textbar\textbf{0.74}
                   
                   & 0.70\textbar0.53\textbar0.70
                   
                   & 0.75\textbar0.67\textbar\textbf{0.77} \\
                   
MLR        & 0.63\textbar0.23\textbar0.63

                   & 0.71\textbar0.38\textbar\textbf{0.73}
                   
                   & 0.49\textbar0.44\textbar\textbf{0.54}
                   
                   & 0.26\textbar0.25\textbar0.26\\
\hline
\end{tabular}\caption{Classification results for OrganCMNIST on Euclidean (left), OT (middle) and UOT (right) embeddings.}
\label{classification_medmnist_organcmnist_raw_hk_wass}
\end{table}

\begin{table}[!htb]
\centering
\scriptsize
\begin{tabular}{|>{\hspace{-3pt}}l<{\hspace{-3pt}}|>{\hspace{-1pt}}c<{\hspace{-1pt}}|>{\hspace{-1pt}}c<{\hspace{-1pt}}|>{\hspace{-1pt}}c<{\hspace{-1pt}}|>{\hspace{-1pt}}c<{\hspace{-1pt}}|}
\hline 
Classifier & MDS & Isomap & t-SNE & Eigenmaps \\
\hline 
    Metric & Euc\textbar OT\textbar UOT & Euc\textbar OT\textbar UOT & Euc\textbar OT\textbar UOT & Euc\textbar OT\textbar UOT
    
    \\ \hline \rule{0pt}{1em}
LDA        & 0.48\textbar0.45\textbar0.45

                   & 0.46\textbar0.43\textbar\textbf{0.51}
                   
                   & $\overline{0.47}$\textbar0.45\textbar0.42
                   
                   & 0.44\textbar0.41\textbar0.44 \\
                   
1NN        & \textbf{0.43}\textbar0.36\textbar0.40

                   & 0.41\textbar0.38\textbar0.43
                   
                   & \textbf{0.44}\textbar0.36\textbar0.40
                   
                   & \textbf{0.44}\textbar0.36\textbar0.39 \\
                   
3NN        & 0.45\textbar0.42\textbar0.43

                   & 0.45\textbar0.41\textbar\textbf{0.48}
                   
                   & 0.44\textbar0.40\textbar0.44
                   
                   & 0.44\textbar0.42\textbar0.44\\
                   
5NN        & \textbf{0.47}\textbar0.44\textbar0.44

                   & 0.44\textbar0.43\textbar\textbf{0.50}
                   
                   & 0.45\textbar0.41\textbar0.44
                   
                   & 0.46\textbar0.42\textbar0.47 \\
                   
SVM(L)   & 0.47\textbar0.46\textbar0.46

                   & 0.47\textbar0.44\textbar\textbf{0.51}
                   
                   & 0.45\textbar0.45\textbar0.44
                   
                   & 0.45\textbar0.44\textbar0.44 \\
                   
SVM(R)   & $\overline{0.47}$\textbar0.47\textbar0.45

                   & 0.46\textbar0.46\textbar\textbf{0.51}
                   
                   & $\overline{0.46}$\textbar0.45\textbar0.41
                   
                   & 0.47\textbar0.45\textbar0.46 \\
                   
RF         & $\overline{0.48}$\textbar0.45\textbar0.45

                   & 0.46\textbar0.45\textbar\textbf{0.51}
                   
                   & \textbf{0.49}\textbar0.40\textbar0.42
                   
                   & 0.47\textbar0.44\textbar0.45 \\
                   
MLR        & \textbf{0.47}\textbar0.56\textbar0.45

                   & 0.46\textbar0.44\textbar\textbf{0.50}
                   
                   & $\overline{0.48}$\textbar0.45\textbar0.44
                   
                   & 0.45\textbar0.44\textbar0.44\\
\hline 
\end{tabular}\caption{Classification results for RetinaMNIST on Euclidean (left), OT (middle) and UOT (right) embeddings.}
\label{classification_medmnist_retinamnist_raw_hk_wass}
\end{table}

All in all, these tables represent 160 classification experiments. The Euclidean metric outperforms both others in 25/160 trials (or 16\% of the time); UOT outperforms both other metrics in 56/160 trials (or 35\% of the time), and OT outperforms the others 3/160 (or 2\% of the time). Interestingly, we find that UOT outperforms OT in 110/160 trials, or 69\% of the time, and UOT outperforms the Euclidean metric in 100 trials (or 45\% of the time). Taken together, these results show that using UOT metrics for neighbor embeddings appear capable of finding more useful low-dimensional embeddings of image data. 

Of particular note, we see that on the MedMNIST datasets, which are more in line with types of data that previous work on OT embeddings have been done, UOT appears to give much better results consistently. Indeed, the Euclidean metric outperforms the rest only 9\% of the time, while UOT outperforms both other metrics 43\% of the time, and UOT outperforms OT 81\% of the time. Taken together, these experiments indicate that the Hellinger--Kantorovich metrics are finding a more meaningful low-dimensional structure in these datasets, especially compared to standard OT. 

\section{Conclusion}   

In this work, we studied how the choice of metric when modeling image data geometry affects task performance in a neighbor embedding framework. In particular, we showed that using the Hellinger--Kantorovich metric from unbalanced optimal transport yields better classification and clustering performance compared to OT-based embeddings a majority of the time, especially on medical imaging benchmarks from the MedMNIST datasets. We also showed that UOT embeddings yield performance improvements over standard Euclidean embeddings some of the time. These results indicate that the constraints of optimal transport, viz. having total mass 1, are too rigid to account for datasets with varying masses and complex structures. We attribute the success of the UOT embeddings to the ability of the HK metrics to create and destroy mass at a distance. It would be interesting to further explore exactly what facets of data geometry are better captured in the HK metrics. Based on our experiments, we suspect that HK metrics are able to account for varied textures within images (evidenced by success on BloodMNIST) as well as variations in overall morphology (as in OrganCMNIST).

\section*{Acknowledgments}
Both authors were partially supported by a Research Enhancement Program grant from the College of Science at the University of Texas at Arlington and by the Army Research Office and was accomplished under Grant
Number W911NF-23-1-0213. The views and conclusions contained in this document are those of the authors and
should not be interpreted as representing the official policies, either expressed or implied, of the Army Research
Office or the U.S. Government. The U.S. Government is authorized to reproduce and distribute reprints for
Government purposes notwithstanding any copyright notation herein.



\bibliographystyle{IEEEbib}

\end{document}